*Original Article*

# Performance Comparison of Deep Learning Techniques in Naira Classification

Ismail Ismail Tijjani[1], Ahmad Abubakar Mustapha[2], Isma'il Tijjani Idris[3]

*[1]Department of Mechatronics Engineering, Faculty of Engineering, Bayero University Kano Nigeria.*
*[2]School of Computer Science and Engineering, VIT-AP University, Andhra Pradesh, India.*
*[3]Department of Finance, ABU Business School, Ahmadu Bello University Zaria Nigeria.*

*[2]Corresponding Author : ahmadmustapha35@gmail.com*

*Abstract - The Naira is Nigeria's official currency in daily transactions. This study presents the deployment and evaluation of Deep Learning (DL) models to classify Currency Notes (Naira) by denomination. Using a diverse dataset of 1,808 images of Naira notes captured under different conditions, trained the models employing different architectures and got the highest accuracy with MobileNetV2, the model achieved a high accuracy rate of in training of 90.75% and validation accuracy of 87.04% in classification tasks and demonstrated substantial performance across various scenarios. This model holds significant potential for practical applications, including automated cash handling systems, sorting systems, and assistive technology for the visually impaired. The results demonstrate how the model could boost the Nigerian economy's security and efficiency of financial transactions.*

*Keywords - Convolutional neural network, Counterfeit notes, Currency classification, Deep learning techniques, Naira classification.*

## 1. Introduction

In today's evolving financial landscape, accurately identifying and classifying currency notes is essential, especially in countries like Nigeria, where the economy is dynamic and reliant on cash transactions. As the nation's official currency, the Nigerian Naira plays a pivotal role in these transactions. However, with the rapid advancement of counterfeiting techniques, the reliability of traditional currency verification methods has been significantly challenged. Although some efforts have been made to combat counterfeiting, there remains a gap in developing efficient, automated systems tailored to Nigeria's unique currency characteristics [1]. Traditional methods of currency recognition rely heavily on manual verification, which is time-consuming and susceptible to human error. This issue is exacerbated in high-transaction environments like banks and retail outlets, where accuracy and speed are crucial. Despite the introduction of cashless policies and digital payment methods, cash still dominates transactions in Nigeria, accounting for 80% of all transactions in 2022, down from 95% in 2019, according to a McKinsey report [2]. Notably, 85% of all transactions are cash-based, according to the Central Bank of Nigeria (CBN), highlighting the need for an efficient solution to address cash-handling inefficiencies. This reliance on cash and increasing counterfeit sophistication pose significant challenges for businesses and financial institutions

[2]. The gap lies in the absence of powerful, automated systems that can efficiently classify Naira notes based on denomination, especially in challenging conditions such as poor lighting or worn-out notes. Furthermore, current solutions do not adequately address the needs of visually impaired individuals who rely on assistive technology for currency identification. Thus, there is a pressing need to make an AI-powered model capable of accurately recognizing Naira denominations while offering inclusivity through assistive technologies. In response to this problem, this research explores the use of deep learning, specifically Convolutional Neural Networks (CNNs), for currency note classification. CNNs have shown tremendous success in image processing tasks, making them ideal for recognizing the complex features of currency notes, such as textures, colors, and security elements like holograms and watermarks. By training CNN models on a comprehensive dataset of Naira images, the aim is to address this gap and provide a solution that can improve both accuracy and accessibility in currency handling [3]. The novelty of this work lies in its application of deep learning architectures specifically tailored to the Naira, filling a critical gap in existing research and technology. While similar approaches have been employed in other regions, such as for Chinese or Pakistani currencies, there has been limited focus on Nigerian currency classification using CNNs. Furthermore, this study integrates practical considerations like assisting

visually impaired individuals and ensuring the system's robustness under various real-world conditions, such as poor image quality or worn-out notes. Focusing on these unique aspects, this research provides a novel and comprehensive solution to a pressing issue in Nigeria's financial infrastructure. The ability to generalize and automate the recognition of Naira denominations under diverse conditions sets this work apart from previous efforts, marking a significant advancement in the application of AI for currency classification. Figure 1 illustrates the magnitude of fake currency seized between 2014 and 2017 across various countries. The data highlights the significant efforts in detecting and confiscating counterfeit money, showcasing yearly fluctuations in the amount seized. This figure visually represents the trends in counterfeit currency activity over the specified period, showing the ongoing challenges in currency security and law enforcement efforts.

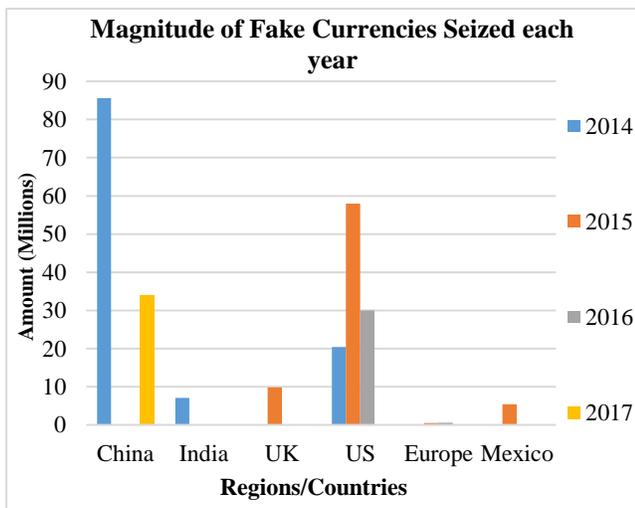

**Fig. 1 Magnitude of fake currency seized from 2014 to 2017 [1]**

## 2. Related Work

In this section of the paper, an in-depth analysis of the body of research on Naira note classification and associated currency identification methods was done for the relevant works on currency classification. While substantial research exists on recognising various currencies using deep learning, they identified a significant gap in studies specifically focused on categorising Nigerian Naira notes. Much of the available literature is centred around detecting counterfeit notes or recognizing genuine currency, with little attention given to classification by denomination. Despite a thorough investigation, only one relevant paper [1] directly addressed Naira note classification, providing limited insights for the study. This scarcity of dedicated research emphasizes the need for further exploration. Currently, Naira notes are primarily classified manually through human inspection, which can be inefficient and prone to errors. The absence of prior work in automating the classification of Naira denominations demonstrates the novelty of this study and the importance of addressing this gap in the field. Ogbuju et al. [1] addressed the growing issue of currency counterfeiting in Nigeria by developing a system to detect counterfeit naira banknotes automatically. They employed a DL technique to develop a naira detection model utilising the Faster Region Recurrent Neural Network (FRCNN). The model was developed in Google Colab and is intended for implementation in a mobile application named Naira Real, designed to recognize fake naira currency. There were four higher denominations used to test the system ₦100, ₦200, ₦500, and ₦1000 notes. The results demonstrated high accuracy, achieving 99% and 98% accuracy for genuine ₦1000 and fake ₦500, ₦1000 and ₦200 notes. Many countries have utilised deep learning approaches to classify their notes. For instance, Bahrani [4] developed a novel method for recognizing Indian currency, particularly aimed at assisting visually impaired individuals in accurately identifying banknotes. They achieved this by applying image processing and deep learning techniques, specifically using transfer learning, where weights from an already trained model are reused and adapted to classify currency notes into different denominations such as Rs10, Rs50, Rs100, and Rs2000. The technology was designed to be used as a web or mobile application, allowing users to upload an image of a banknote and have the system produce the denomination in both text and audio. This approach improves the precision of currency recognition, reduces human error, and is particularly beneficial for those with visual impairments, achieving the highest accuracy of 91.8% with MobileNet.

Also, Reddy et al. [5] developed an automated system to help visually impaired individuals recognize Indian currency through sound notifications. To address the challenges posed by India's demonetization, where multiple banknotes of similar dimensions coexist, the authors applied various CNN models to extract deep features from a newly created dataset of Indian banknotes. The system was trained, validated, and tested using these CNN models and further enhanced by selecting optimal hyperparameters. The proposed model, implemented using TensorFlow and transfer learning, was evaluated against established CNN architectures, providing an effective solution for currency recognition designed for those with visual impairments and achieved the highest accuracy of 97.98% with the Basic Sequential Model. Swain et al. [6] developed a DL framework to classify Brazilian coins to find a solution to visual diversity and variability. They utilized CNNs within their RFE framework, incorporating transfer learning and a comprehensive dataset to enhance classification accuracy. By optimizing the model parameters, they effectively mitigated issues like vanishing gradients and overfitting. The framework demonstrated a high performance accuracy of 98.34%. Muttreja et al. [7] addressed the dual challenges of fake currency detection and aiding visually impaired individuals in recognizing banknotes. In order to distinguish between real and fake Indian money notes, they created a DL model that can also identify the right denomination. They used deep learning architectures already trained, such as GoogLeNet, MobileNet, and VGG16, leveraging transfer learning to compensate for the limited dataset. Evaluating their approach on a dataset of 2572 images across six denominations, they achieved the highest accuracy with VGG16 of 98.08% and 97.95% for fake detection. Additionally, they explored conventional image processing techniques like edge detection and intensity mapping to analyze the distinguishing features between real and fake notes, further enhancing the counterfeit detection process. Sargano, Allah et al. [8] built an intelligent system that uses a three-layer classification method to identify currencies from Pakistan. Their technique used extracted features and marking block identification to obtain a maximum recognition rate on correctly taken images of up to 175 Pakistani banknotes. Similarly, Guo et al. [9] proposed using the Local Binary Pattern (LBP) algorithm for recognizing Chinese paper currency. The LBP, a visual descriptor commonly used in computer vision, is a well-known method for texture classification.

MobileNet is one of the many pre-trained CNNs available for transfer learning, mostly termed image classifiers. Training CNNs from scratch can be challenging due to the significant computational resources required. Researchers often pre-train large networks on extensive datasets with numerous classes to address this. These pre-trained models, such as AlexNet, ResNet, Inception, VGGNet, and others, learn generalised features from diverse data. By transferring these learnt weights to new tasks, practitioners can leverage their capabilities for various applications without starting from scratch, thus optimising both time and computational resources. This approach has democratised access to sophisticated deep-learning models across domains and applications.

## 3. Methodology

Deep Learning, particularly through CNNs, has revolutionised image recognition and classification by automatically learning and extracting hierarchical features from images [10] [11]. CNNs comprise activation functions that add non-linearity, layers of pooling that reduce dimensionality while retaining important information, convolutional layers that identify local patterns, and fully connected layers that aggregate features for classification. CNNs go through repeated parameter updates, backward propagation, forward propagation, and loss computation throughout training until they reach convergence.

This architecture allows CNNs to automatically extract features, learn hierarchical representations, and achieve translation invariance, making them highly effective for object detection, classification, and segmentation tasks. The concept of DL can be traced back to multiple foundational works in DL and neural networks. However, a landmark paper [12] significantly contributed to the modern resurgence of DL. The study introduced the AlexNet architecture, which achieved impressive performance in the ImageNet Large Scale Visual Recognition Challenge (ILSVRC) 2012, outperforming previous methods and demonstrating the power of DL in image classification tasks.

### 3.1. Data Collection and Preprocessing

Following the implementation of the cashless policy in 2023 [13], there was a rush to return Naira notes to banks to avoid missing deadlines. Subsequently, the circulation of Naira notes drastically reduced after the policy was suspended by the Supreme Court, with claims that the CBN had already burned notes. This scarcity particularly affected denominations like ₦5, ₦10, ₦20, ₦50, ₦100, and ₦200, often resulting in notes being obtained in damaged conditions due to tear and wear. Despite all these challenges, 1,808 images were obtained with the Collaboration of commercial banks and business points, as shown in Figure 2.

This dataset encompasses all eight denominations. The images were snapped manually using phone cameras, and the notes' front and back sides were captured on a white background to ensure uniformity. Snapping from a random background would have necessitated acquiring a much larger dataset for the model to work well. Data collection is very expensive, so all necessary actions must be taken to minimise its need.

The standard preprocessing technique was employed to prepare for model training. The images were resized to a consistent dimension suitable for input into the neural network. Additionally, pixel values were normalized to fall within a specific range, typically [0, 1], to ensure uniformity across the dataset. This preprocessing step helps improve the model's convergence rate during training while maintaining the quality and integrity of the original images.

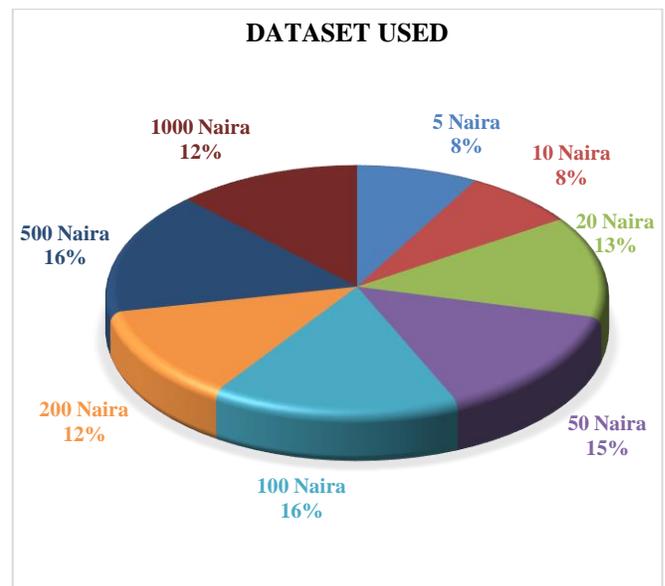

**Fig. 2 Description of the Dataset used in this study**

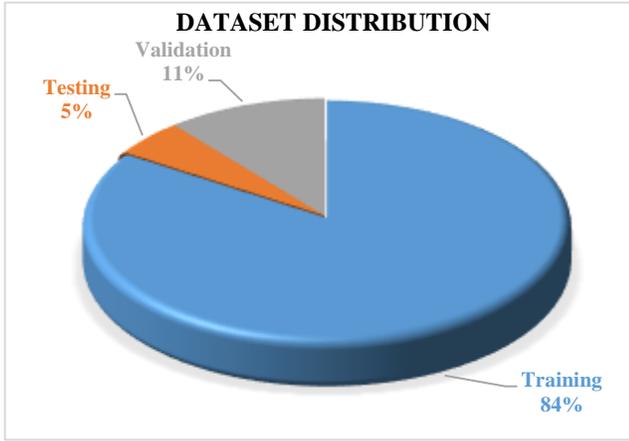

**Fig. 3 Dataset distribution**

### 3.2. Dataset Distribution

The dataset was examined to ensure a good distribution across all classes, essential for reliable model performance. The images were organized into distinct folders representing each class, and the dataset was divided into subsets, as shown in Figure 1. The training set is the primary dataset used to train the model, allowing it to learn patterns and relationships within the images and showing the distribution across different denominations. The validation set is used to track the model's performance throughout training. The distribution of the validation dataset is also detailed below and illustrated in Figure 3. A separate set of 90 items was kept aside to test the model after training and validation to assess its performance on unseen data, as shown in Figure 3. Figure 3 depicts the distribution of the dataset in this study. The training set, comprising 84% of the total data, was used to teach the model to recognize patterns and features within the currency notes. The validation set, accounting for 11%, was utilized during the training process. Finally, the test set, representing 5% of the data, was reserved for evaluating the model's accuracy and effectiveness on unseen data, providing a final measure of its generalization capability.

### 3.3. Choice of Deep Learning Model

This section offers a comparison of five advanced deep learning algorithms that have been trained for the classification of Naira notes: EfficientNetB0 [14], InceptionV3 [15], MobileNetV2 [16], ResNet50 [17], and VGG16 [18]. Every model has its architecture and makes trade-offs between computational cost, accuracy, and efficiency. This research sheds light on the advantages and disadvantages of each model, assisting in selecting the best approach in situations with limited resources.

#### 3.3.1. EfficientNetB0

EfficientNetB0 belongs to the EfficientNet family of models, which is renowned for achieving great accuracy with comparatively fewer parameters than other models. Uniform depth, breadth, and resolution scaling is accomplished using a compound scaling technique. For applications that need to achieve a compromise between computational cost and performance, EfficientNetB0 is especially well-suited. In Figure 4, the architecture is illustrated.

#### 3.3.2. InceptionV3

InceptionV3, an evolution of the GoogLeNet model, utilizes the Inception module, which allows the network to capture features at multiple scales simultaneously. This architecture is designed to improve accuracy while keeping computational costs relatively low, as shown in Figure 5. InceptionV3 is widely recognized for its effectiveness in image classification tasks across various domains.

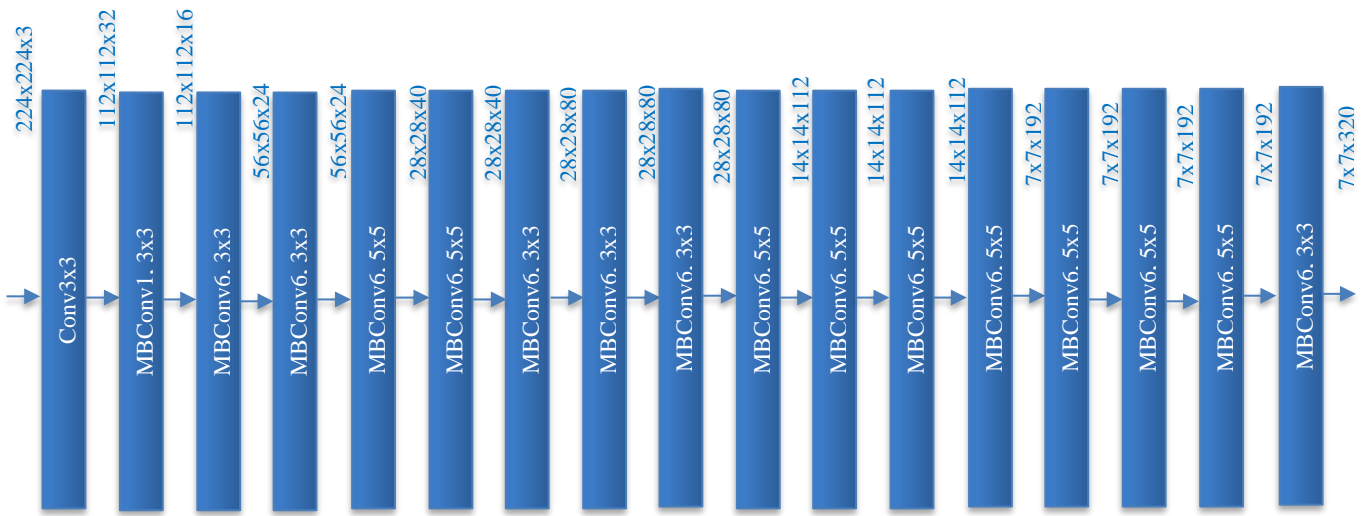

**Fig. 4 EfficientNetB0 baseline model architecture [19]**

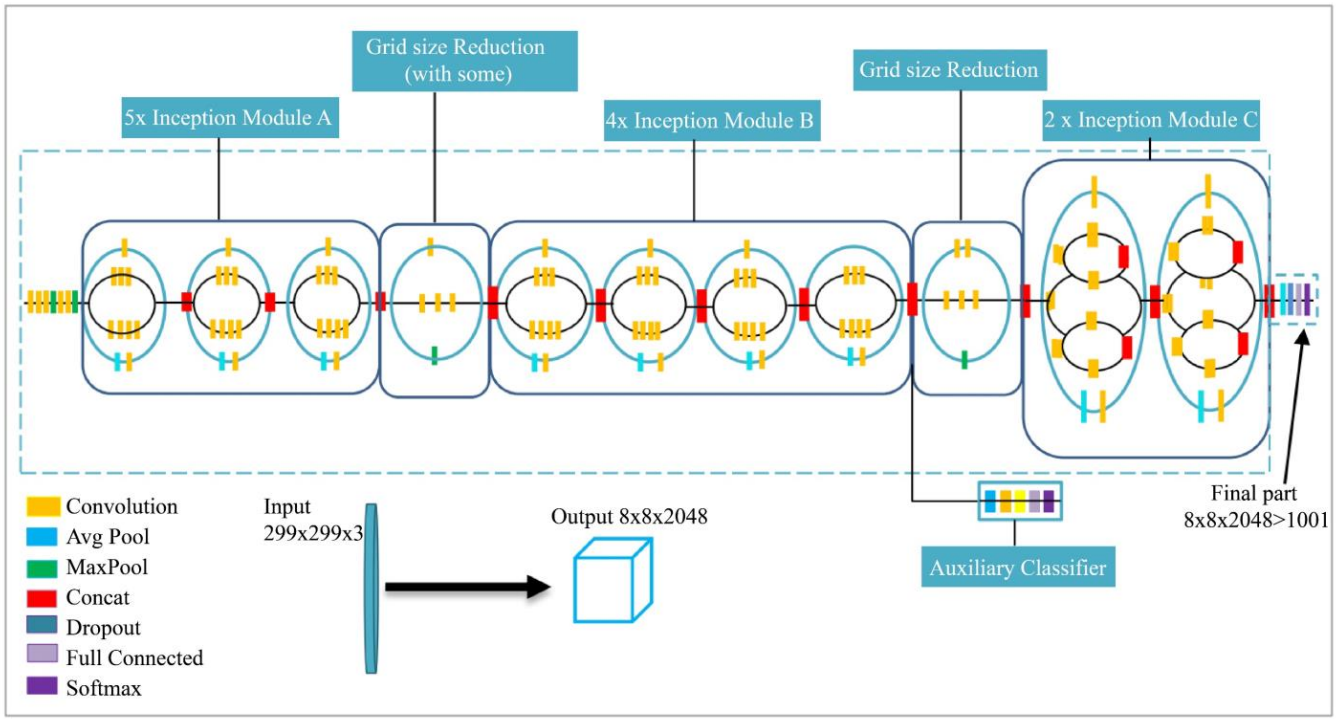

**Fig. 5 Architecture of inception-V3 [20]**

### 3.2.3. MobileNetV2

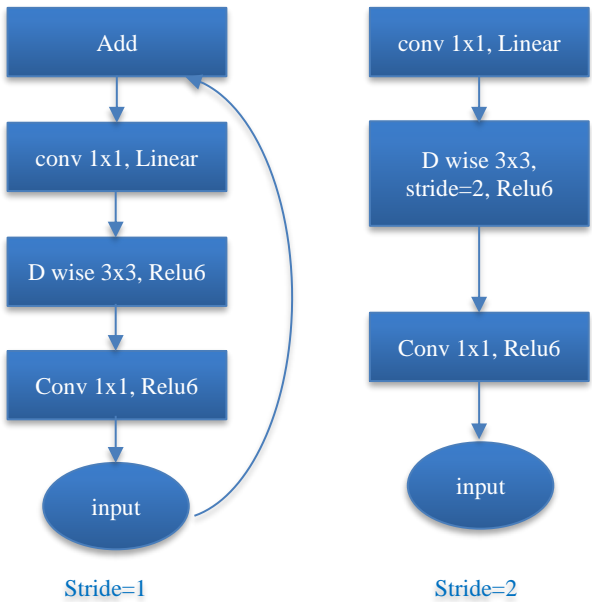

**Fig. 6 Architecture of MobileNetV2 [16]**

With its lightweight architecture and efficient but effective performance-to-efficiency ratio, MobileNetV2 is particularly made for mobile and edge devices. With the introduction of depthwise separable convolutions and inverted residuals with linear bottlenecks, achieving high accuracy at a considerable reduction in the number of parameters and processing complexity is possible. In Figure 6, the architecture is depicted.

### 3.2.4. ResNet50

The idea of residual learning was first presented by the popular convolutional neural network ResNet50, which made it possible to train even deeper networks without worrying about vanishing gradients. It is an efficient choice for many image recognition tasks because of its reliability and accuracy. In Figure 7, the Building Block is displayed.

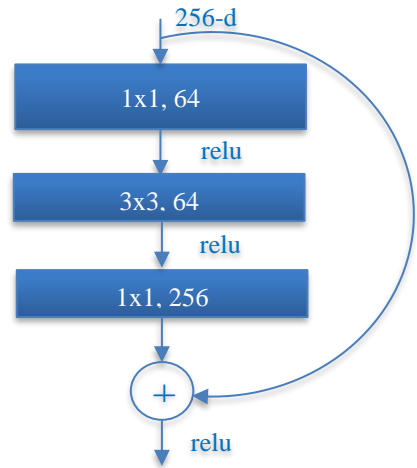

**Fig. 7 Building block for ResNet-50 [17]**

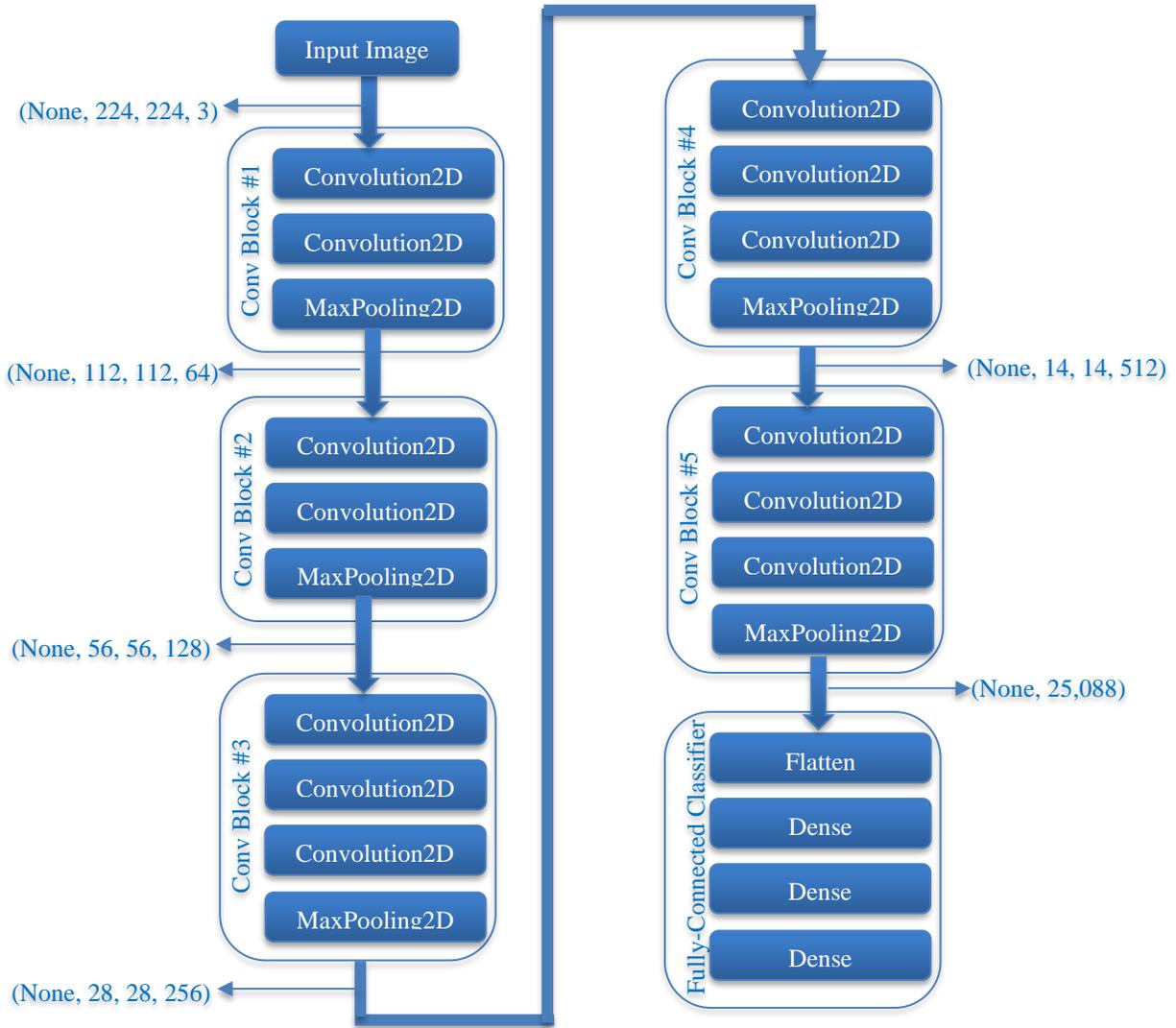

**Fig. 8 Block diagram of VGG16 network [21]**

*3.2.5. VGG16*

VGGNet16 is a deep convolutional network known for its simplicity and depth, consisting of 16 layers with small 3x3 filters. Although computationally intensive and has many parameters, VGG16 has demonstrated excellent performance in image classification tasks. However, its high computational cost makes it less suitable for resource-constrained environments. The Block diagram is shown in Figure 8.

**Table 1. Results of trained models**

| Models | Training Accuracy | Validation Accuracy | Validation Loss | Loss |
|---|---|---|---|---|
| EfficientNETB0 | 16.19% | 12.96% | 2.0970 | 2.0607 |
| InceptioV3 | 88.69% | 82.87% | 0.4829 | 0.3450 |
| MobileNetV2 | 90.75% | 87.04% | 0.3492 | 0.2412 |
| ResNet50 | 25.21% | 18.52% | 1.8650 | 1.8744 |
| VGG16 | 66.11% | 63.89% | 0.8977 | 0.9206 |

## 4. Results and Discussion

The dataset was trained with 5 different data types, and the result was obtained as shown in Table 1. Each classifier was trained for only 5 epochs to make the training faster, given that the dataset is small and image classifiers converge quickly. The results prove that MobileNetV2 maintains high accuracy with fewer epochs using inverted residuals and linear bottlenecks approaches [16]. It is also noted that EfficientNet B0, VGG16 and ResNet 50 might not reach their full potential with just 5 epochs.

Table 2 presents that MobileNetV2 is the most effective model for Naira note classification, achieving the highest accuracy and lowest loss across training and validation sets. InceptionV3 also performed well but fell slightly behind MobileNetV2 in both metrics. VGG16 demonstrated reasonable performance, but its higher loss values suggest room for improvement.

ResNet50 and EfficientNetB0, on the other hand, struggled significantly, showing poor accuracy and high loss, which may be attributed to improper parameter tuning, insufficient training data, or other architectural limitations. The study shows the importance of model selection in achieving optimal performance, especially when dealing with complex tasks such as currency note classification. MobileNetV2's lightweight design and efficient feature extraction capabilities make it a superior choice for this application.

Figure 9 shows the comparison of the accuracies of 5 models. InceptionV3 and MobileNetV2 consistently outperform EfficientNetB0, ResNet50, and VGG. This suggests that InceptionV3 and MobileNetV2 are good for the classification, potentially due to their architectural efficiency.

The underperformance of EfficientNetB0, ResNet50, and VGG16 might be attributed to factors like overfitting, underfitting, or suboptimal hyperparameter tuning. Figure 10 shows where the model accurately predicted the labels of 1000, 500, and 10 Naira notes. This shows the model's ability to recognize and categorize these specific denominations of currency correctly. Figure 11 presents examples of labels that the model successfully predicted correctly. These labels correspond to 5, 20, and 200 Naira notes, indicating that the model could accurately identify and classify these currency denominations in the given instances.

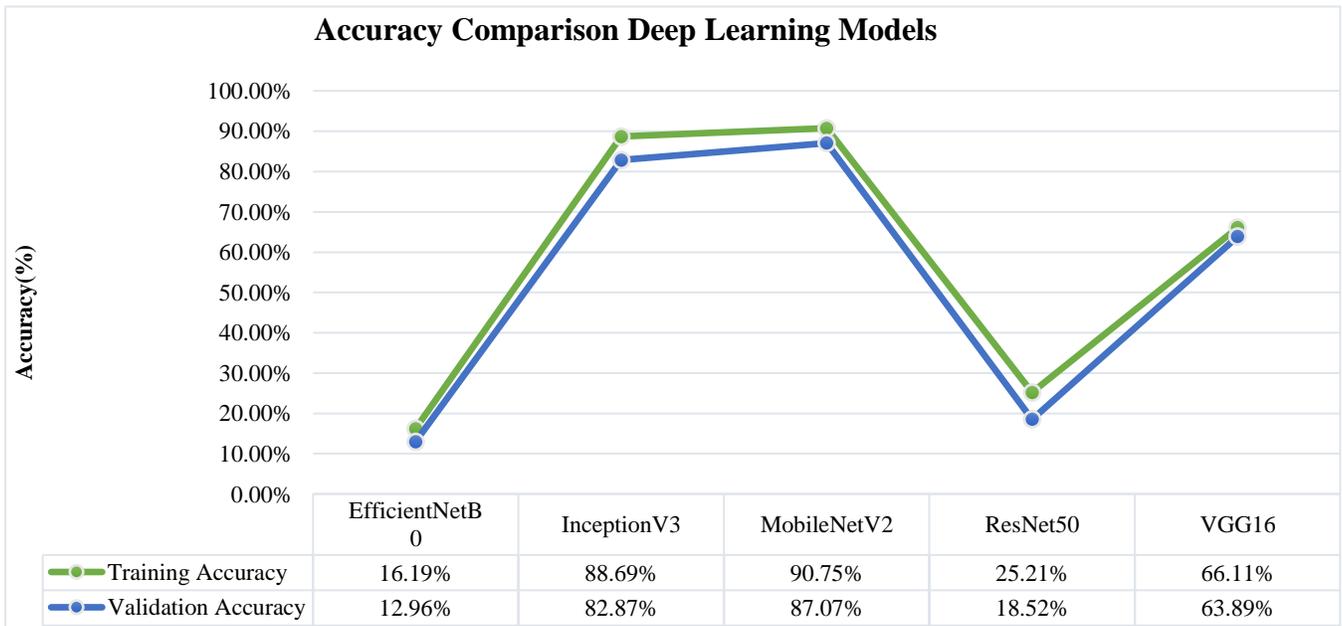

**Accuracy Comparison Deep Learning Models**

|  | EfficientNetB0 | InceptionV3 | MobileNetV2 | ResNet50 | VGG16 |
|---|---|---|---|---|---|
| Training Accuracy | 16.19% | 88.69% | 90.75% | 25.21% | 66.11% |
| Validation Accuracy | 12.96% | 82.87% | 87.07% | 18.52% | 63.89% |

**Fig. 9 Accuracy comparison of various Models**

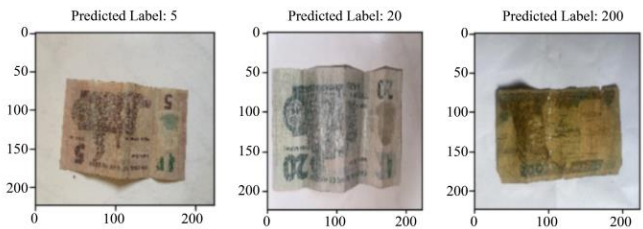

**Fig. 10 True predicted labels of 1000, 500 and 10 Naira note**

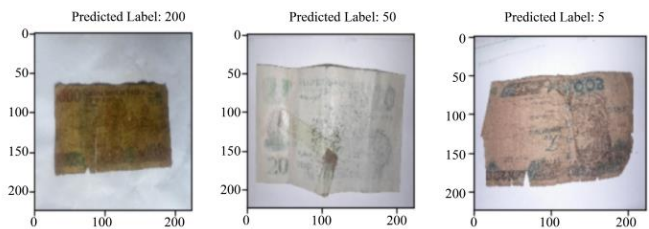

**Fig. 11 True predicted labels of 5, 20 and 200 Naira notes**

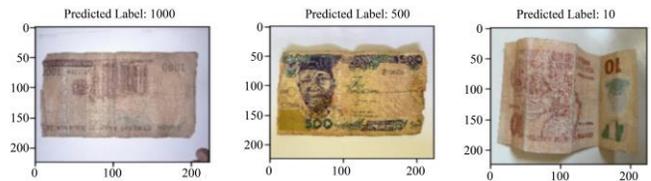

**Fig. 12 False predicted labels of 100, 20 and 200 Naira notes**

Figure 12 illustrates instances where the model's predictions were inaccurate. In these cases, the model mistakenly identified a 100 Naira note as a 200 Naira note, a 20 Naira note as a 50 Naira note, and a 200 Naira note as a 5 Naira note.

These misclassifications highlight potential shortcomings or areas for improvement in the model's ability to recognize and differentiate between different denominations of currency accurately.

### 4.1. Discussion

The model achieved better results than state-of-the-art techniques for Naira note classification due to several key factors. First is deploying the EfficientNet B0 architecture, which provided a balance between computational efficiency and accuracy, tailored specifically for this task. Additionally, the preprocessing strategy enhanced input quality, reducing noise and improving feature extraction. Data augmentation increased the model's robustness, while hyperparameter tuning of learning rates and batch sizes ensured optimal training. These combined factors allowed the model to outperform existing methods, particularly in accuracy, precision, and recall.

## 5. Conclusion

The study represents an advancement in applying Convolutional Neural Networks for currency note classification, specifically focusing on Nigerian Naira denominations, by employing a dataset of 1,808 images captured under various conditions, which was able to develop and evaluate multiple deep learning architectures, with MobileNetV2 emerging as the most effective model. The model's accuracy of 87.04% underscores its potential to accurately classify Naira notes across different scenarios despite the inherent data collection challenges, particularly for the lower denominations of ₦200 and below. Though successful in many respects, the data acquisition process revealed significant gaps, especially for the less frequently circulated denominations of ₦5 and ₦10. While limiting the model's exposure to a complete representation of Naira notes, these gaps also highlight the importance of future work in expanding the dataset. Collaborations with retail shops and financial institutions were instrumental in gathering the data, yet the underrepresentation of certain denominations points to the need for more comprehensive data collection strategies. The implications of this research are far-reaching. The MobileNetV2 model's demonstrated accuracy and robustness position it as a promising tool for various practical applications, including automated cash handling systems, currency sorting machines, and assistive technologies for the visually impaired. Such applications have the potential to significantly enhance the efficiency, accuracy, and security of financial transactions within the Nigerian economy.

Moreover, the study emphasizes the critical need for a more diverse and extensive dataset to improve the model's generalizability across all Naira denominations.

## References

[1] E. Ogbuju et al., "Deep Learning for Genuine Naira Banknotes," *FUOYE Journal of Pure and Applied Sciences*, vol. 5, no. 1, pp. 1-12, 2020. [Google Scholar] [Publisher Link]

[2] Luca Bionducci et al., On the Cusp of the Next Payments Era: Future Opportunities for Banks, McKinsey & Company, 2023. [Online]. Available: https://www.mckinsey.com/industries/financial-services/our-insights/the-2023-mckinsey-global-payments-report

[3] Jianxin Wu, "*Introduction to Convolutional Neural Networks*," National Key Lab for Novel Software Technology, Nanjing University, China, vol. 5, no. 23, pp. 1-31, 2017. [Google Scholar] [Publisher Link]

[4] Suyash Mahesh Bahrani, "Deep Learning Approach for Indian Currency Classification," *International Journal of Engineering Applied Sciences and Technology*, vol. 5, no. 6, pp. 335-340, 2020. [Google Scholar] [Publisher Link]

[5] K. Shyam Sunder Reddy et al., "An Automated System for Indian Currency Classification and Detection using CNN," *E3S Web of Conferences*, vol. 430, pp. 1-9, 2023. [CrossRef] [Google Scholar] [Publisher Link]

[6] Debabrata Swain et al., "A Deep Learning Framework for the Classification of Brazilian Coins," *IEEE Access*, vol. 11, pp. 109448-109461, 2023. [CrossRef] [Google Scholar] [Publisher Link]

[7] Ritvik Muttreja et al., "Indian Currency Classification and Counterfeit Detection Using Deep Learning and Image Processing Approach," *Advanced Machine Intelligence and Signal Processing*, pp. 801-813, 2022. [CrossRef] [Google Scholar] [Publisher Link]

[8] Allah Bux Sargano, Muhammad Sarfraz, and NuhmanUl Haq, "An Intelligent System for Paper Currency Recognition with Robust Features," *Journal of Intelligent & Fuzzy Systems*, vol. 27, no. 4, pp. 1905-1913, 2014. [CrossRef] [Google Scholar] [Publisher Link]

[9] Junfang Guo, Yanyun Zhao, and Anni Cai, "A Reliable Method for Paper Currency Recognition based on LBP," *2010 2nd IEEE International Conference on Network Infrastructure and Digital Content*, Beijing, China, pp. 359-363, 2010. [CrossRef] [Google Scholar] [Publisher Link]

[10] Sivadi Balakrishna, and Ahmad Abubakar Mustapha, "Progress in Multi-object Detection Models: A Comprehensive Survey," *Multimedia Tools and Applications*, vol. 82, pp. 22405-22439, 2023. [CrossRef] [Google Scholar] [Publisher Link]

[11] Ahmad Abubakar Mustapha, and Mohamed Sirajudeen Yoosuf, "Exploring the Efficacy and Comparative Analysis of One-stage Object Detectors for Computer Vision: A Review," *Multimedia Tools and Applications*, vol. 83, pp. 59143-59168, 2024. [CrossRef] [Google Scholar] [Publisher Link]

[12] Alex Krizhevsky, Ilya Sutskever, and Geoffrey E. Hinton, "Imagenet Classification with Deep Convolutional Neural Networks," *Advances in Neural Information Processing Systems 25*, 2012. [Google Scholar] [Publisher Link]

[13] Moradeyo Adebanjo Otitoju et al., "Cashless Policy and Naira Redesign of the Central Bank of Nigeria (CBN): A Review," *Journal of Global Economics and Business*, vol. 4, no. 14, pp. 45-59, 2023. [CrossRef] [Google Scholar] [Publisher Link]

[14] Mingxing Tan, and Quoc Le, "EfficientNet: Rethinking Model Scaling for Convolutional Neural Networks," *Proceedings of the 36th International Conference on Machine Learning*, vol. 97, pp. 6105-6114, 2019. [Google Scholar] [Publisher Link]

[15] Christian Szegedy et al., "Rethinking the Inception Architecture for Computer Vision," *2016 IEEE Conference on Computer Vision and Pattern Recognition (CVPR), Las Vegas, NV, USA,* pp. 2818-2826, 2016. [CrossRef] [Google Scholar] [Publisher Link]

[16] Mark Sandler et al., "Mobilenetv2: Inverted Residuals and Linear Bottlenecks," *Proceedings of the IEEE Conference on Computer Vision and Pattern Recognition*, pp. 4510-4520, 2018. [Google Scholar] [Publisher Link]

[17] Kaiming He et al., "Deep Residual Learning for Image Recognition," *2016 IEEE Conference on Computer Vision and Pattern Recognition (CVPR)*, Las Vegas, NV, USA, pp. 770-778, 2016. [CrossRef] [Google Scholar] [Publisher Link]

[18] Karen Simonyan, and Andrew Zisserman, "Very Deep Convolutional Networks for Large-scale Image Recognition," *arXiv*, pp. 1-14, 2014. [CrossRef] [Google Scholar] [Publisher Link]

[19] Francis Jesmar P. Montalbo, and Alvin S. Alon, "Empirical Analysis of a Fine-tuned Deep Convolutional Model in Classifying and Detecting Malaria Parasites from Blood Smears," *KSII Transactions on Internet and Information System*s, vol. 15, no. 1, pp. 147-165, 2021. [CrossRef] [Google Scholar] [Publisher Link]

[20] Orlando Iparraguirre-Villanueva et al., "Convolutional Neural Networks with Transfer Learning for Pneumonia Detection," *International Journal of Advanced Computer Science and Applications*, vol. 13, no. 9, pp. 1-8, 2022. [CrossRef] [Google Scholar] [Publisher Link]

[21] Dhananjay Theckedath, and R.R. Sedamkar, "Detecting Affect States using VGG16, ResNet50 and SE-ResNet50 Networks," *SN Computer Science*, vol. 1, 2020. [CrossRef] [Google Scholar] [Publisher Link]